\documentclass[review,authoryear]{elsarticle}

\graphicspath{ {./figs/} }
\usepackage[table, dvipsnames]{xcolor} 
\usepackage{graphicx}

\usepackage{amsmath, amssymb, amsthm, amsfonts, bm}

\usepackage{booktabs}
\usepackage{multirow}
\usepackage{tabularx}
\usepackage{threeparttable}
\usepackage{ragged2e}
\usepackage{cite}

\usepackage{natbib}
\let\cite\citep

\usepackage[ruled,linesnumbered]{algorithm2e}
\SetCommentSty{textcolor{gray!60}}

\usepackage{enumitem}
\usepackage{pifont}

\usepackage{cite}
\usepackage{hyperref} 

\biboptions{authoryear}
\definecolor{mydarkblue}{rgb}{0, 0.08, 0.45}

\title{CoFusion: Multispectral and Hyperspectral Image Fusion via Spectral Coordinate Attention}

\author[label1]{Baisong Li}
\ead{libaisonm@outlook.com}
\address[label1]{School of Integrated Circuits, Tsinghua University, Beijing 100084, China}

\begin{document}
\begin{frontmatter}
\begin{abstract}
Multispectral and Hyperspectral Image Fusion (MHIF) aims to reconstruct high-resolution images by integrating low-resolution hyperspectral images (LRHSI) and high-resolution multispectral images (HRMSI). However, existing methods face limitations in modeling cross-scale interactions and spatial-spectral collaboration, making it difficult to achieve an optimal trade-off between spatial detail enhancement and spectral fidelity.
To address this challenge, we propose CoFusion: a unified spatial-spectral collaborative fusion framework that explicitly models cross-scale and cross-modal dependencies. Specifically, a Multi-Scale Generator (MSG) is designed to construct a three-level pyramidal architecture, enabling the effective integration of global semantics and local details. Within each scale, a dual-branch strategy is employed: the Spatial Coordinate-Aware Mixing module (SpaCAM) is utilized to capture multi-scale spatial contexts, while the Spectral Coordinate-Aware Mixing module (SpeCAM) enhances spectral representations through frequency decomposition and coordinate mixing. Furthermore, we introduce the Spatial-Spectral Cross-Fusion Module (SSCFM) to perform dynamic cross-modal alignment and complementary feature fusion.
Extensive experiments on multiple benchmark datasets demonstrate that CoFusion consistently outperforms state-of-the-art methods, achieving superior performance in both spatial reconstruction and spectral consistency.
\end{abstract}



\end{frontmatter}

\section{Introduction}
\label{sec:introduction}

Multispectral and Hyperspectral Image Fusion (MHIF) aims to reconstruct a high-resolution hyperspectral image (HRHSI) by integrating the rich spectral information of a low-resolution hyperspectral image (LRHSI) with the fine spatial details of a high-resolution multispectral image (HRMSI). This task effectively alleviates the inherent trade-off between spatial and spectral resolutions in remote sensing systems, providing critical data support for a wide range of applications such as land-cover classification~\cite{class_01,class_02,class_03,class_04}, environmental monitoring, and precision agriculture~\cite{agriculture}. However, MHIF is essentially a highly ill-posed inverse problem. Its core challenge lies in achieving precise spatial detail reconstruction while maintaining spectral fidelity under significant cross-scale and cross-modal discrepancies.

Early research primarily relied on model-based methods, including component substitution (CS)~\cite{PCA,laben2000process}, multi-resolution analysis (MRA)~\cite{MRA_01,MRA_02,MRA_03}, and variational optimization (VO)~\cite{vo_01,vo_02} approaches. CS-based methods inject spatial details directly via transform-domain substitution; however, they often lead to severe spectral distortion due to oversimplified assumptions regarding spectral independence. MRA-based methods mitigate this issue by utilizing multi-scale decomposition, yet they remain sensitive to filter design and are prone to spatial artifacts such as ringing and blurring. VO-based methods introduce hand-crafted priors to regularize the solution space; nevertheless, their limited representational capacity and high computational costs restrict their applicability in complex real-world scenarios.

With the rapid development of deep learning, data-driven methods have become the mainstream paradigm for MHIF. Existing approaches are mainly based on convolutional neural network (CNN)~\cite{HSRnet,3DT-Net,MHFnet,zhu2024implicit} and Transformer~\cite{Fusformer,ESSAformer} architectures. CNN-based models effectively capture local spatial structures through hierarchical feature extraction but lack the ability to model long-range dependencies and global contextual consistency. In contrast, Transformers utilize self-attention mechanisms to establish global feature interactions; however, their quadratic computational complexity poses significant challenges for high-dimensional hyperspectral data. Furthermore, the lack of strong inductive biases for local structures often results in over-smoothing or structural artifacts.

To overcome these limitations, recent works have explored hybrid architectures and dynamic modeling strategies~\cite{ESSAformer}. Although these methods have improved representational capacity to some extent, they still struggle to resolve the fundamental contradiction between spatial detail enhancement and spectral consistency. In particular, most existing methods employ homogeneous or static processing strategies across different regions and scales, lacking adaptive mechanisms to handle diverse spatial structures and spectral characteristics. Consequently, they tend to produce over-smoothing in edge-rich regions while introducing noise in homogeneous areas.

Overall, existing methods exhibit the following limitations:

\begin{itemize}[noitemsep, topsep=0pt]
    \item Inadequate Cross-scale Modeling: The vast resolution difference between LRHSI and HRMSI requires effective interaction between global semantics and local details. However, most methods process different scales independently, leading to sub-optimal fusion performance.
    
    \item Decoupled Spatial-Spectral Modeling: Spatial and spectral features are typically extracted separately, lacking sufficient cross-modal interaction, which results in spatial distortion or spectral inconsistency.
    
    \item Lack of Adaptive Feature Fusion: Existing fusion strategies are generally static and fail to dynamically balance spatial enhancement and spectral preservation under diverse scene conditions.
\end{itemize}

To address these challenges, we propose CoFusion: a unified spatial-spectral collaborative fusion framework that explicitly models cross-scale and cross-modal interactions. The core idea of CoFusion is to achieve joint optimization of spatial and spectral representations through multi-scale hierarchical modeling and adaptive feature fusion.

Specifically, we first design a Multi-Scale Generator (MSG) to construct a three-level pyramidal architecture, enabling effective interaction between global semantic information and local spatial details. Within each scale, we introduce a dual-branch collaborative mechanism: the Spatial Coordinate-Aware Mixing (SpaCAM) module captures multi-scale spatial contexts via gated dilated convolutions, while the Spectral Coordinate-Aware Mixing (SpeCAM) module exploits intrinsic spectral correlations through frequency decomposition and coordinate mixing. Furthermore, to bridge the gap between spatial and spectral representations, we propose the Spatial-Spectral Cross-Fusion Module (SSCFM), which utilizes large-kernel attention and channel excitation to perform dynamic feature alignment and interaction, effectively reducing redundancy while preserving complementary information.

Extensive experiments on multiple benchmark datasets demonstrate that CoFusion consistently outperforms state-of-the-art methods, achieving superior performance in both spatial detail reconstruction and spectral fidelity preservation.

The main contributions of this paper are summarized as follows:
\begin{itemize}
    \item We propose CoFusion: a unified spatial-spectral collaborative fusion framework that explicitly models cross-scale and cross-modal interactions, effectively alleviating the inherent trade-off between spatial detail reconstruction and spectral fidelity preservation in MHIF.

    \item We design the MSG with a pyramidal hierarchical structure, enabling progressive interaction between global semantic representations and local spatial details, thereby improving the alignment and consistency of cross-scale features.

    \item We introduce a spatial-spectral collaborative modeling mechanism consisting of SpaCAM and SpeCAM, which jointly model spatial contexts and spectral dependencies through coordinate-aware mixing and frequency-aware spectral decomposition, enhancing feature representation from both spatial and spectral perspectives.

    \item We propose the SSCFM to achieve adaptive feature alignment and dynamic interaction between spatial and spectral representations, effectively reducing redundancy while preserving complementary information.
\end{itemize}

The remainder of this paper is organized as follows: Section~\ref{sec:related_work} reviews related work in hyperspectral image fusion, including CS, MRA, VO methods, and deep learning-based approaches. Section~\ref{sec:method} provides a detailed description of the proposed CoFusion framework, including MSG, SpaCAM, SpeCAM, and SSCFM. Section~\ref{sec:experiments} describes the experimental setup, datasets, and evaluation metrics, followed by a comprehensive comparison with state-of-the-art methods and ablation studies. Finally, Section~\ref{sec:conclusion} concludes the paper and discusses potential future research directions.

\section{Related Work}
\label{sec:related_work}

\subsection{Review of MHIF Methods}
Multispectral and Hyperspectral Image Fusion (MHIF) has emerged as a fundamental task in the field of remote sensing, aiming to reconstruct high-fidelity hyperspectral images by integrating high-resolution multispectral images (HRMSI) to compensate for the spatial sparsity of low-resolution hyperspectral images (LRHSI). Existing methods can be broadly categorized into traditional model-based methods and deep learning-based methods.

\subsubsection{Traditional MHIF Methods}
Traditional MHIF methods formulate the fusion process through explicit mathematical modeling and are generally classified into component substitution (CS), multi-resolution analysis (MRA), and variational optimization (VO) methods. CS-based methods project multispectral data into a transform domain and replace specific components with spatial details extracted from the HRMSI. Representative methods such as PCA~\cite{PCA}, IHS~\cite{laben2000process}, and BDSD~\cite{BDSD} are computationally efficient but often produce spectral distortion due to the global substitution operation. MRA-based methods extract spatial details across multiple scales using multi-resolution decomposition techniques, such as SFIM~\cite{SFIM}, MTF-GLP~\cite{MTF-GLP}, AWLP~\cite{AWLP}, and DWT~\cite{DWT}. Although these methods generally improve spectral preservation, they may still introduce spatial artifacts and aliasing effects. VO-based methods~\cite{vo_01,vo_02} model MHIF as a constrained optimization problem by introducing spatial and spectral priors. While these methods provide strong theoretical guarantees, their iterative optimization processes typically result in high computational costs, limiting their scalability for large-scale applications.

\subsubsection{Deep Learning-based MHIF Methods}
Deep learning has significantly advanced MHIF by replacing hand-crafted priors with data-driven hierarchical representations. Early CNN-based methods, such as MHFnet~\cite{MHFnet} and HSRnet~\cite{HSRnet}, effectively capture local spatial structures but face limitations in modeling long-range spectral dependencies due to their restricted receptive fields.

To address this limitation, Transformer-based models, such as Fusformer~\cite{Fusformer}, introduced self-attention mechanisms to achieve global context modeling. However, the quadratic complexity of self-attention poses computational challenges. Subsequent methods, including PSRT~\cite{PSRT}, U2Net~\cite{U2Net}, and 3DT-Net~\cite{3DT-Net}, explored more efficient architectures to balance performance and complexity.

Recently, hybrid and dynamic fusion strategies have gained increasing attention. SMGU-Net~\cite{SMGU-Net} combines optimization-inspired modeling with deep networks but still struggles with complex cross-modal non-linear interactions. Dynamic approaches, such as FMPM-DNet~\cite{FMPM-DNet} and BUGPan~\cite{BUGPan}, further improve adaptivity through modulation and uncertainty-aware mechanisms. Despite these advancements, existing methods still face challenges in simultaneously maintaining local sharpness, global structural consistency, and spectral fidelity.

\subsection{Attention-based Fusion Mechanisms}
Attention mechanisms have been widely adopted in MHIF to enhance feature representation and facilitate cross-modal information integration. Existing methods typically employ channel, spatial, or hybrid attention to emphasize meaningful features while suppressing redundancy. Channel attention models improve spectral consistency by capturing inter-band dependencies, while spatial attention focuses on salient regions to enhance the preservation of structural details. Recently, Large Kernel Attention (LKA)~\cite{lka1,lka2,lka3} and non-local operations~\cite{liu2021Swin,hu2019local,liu2021swinv2,li2026pif} have been introduced to better model long-range spatial dependencies.

Despite their effectiveness, existing attention-based fusion strategies still exhibit two primary limitations. First, attention is often applied within a single domain, which limits spatial-spectral interaction modeling. Second, many fusion schemes are static or weakly coupled, restricting the adaptive alignment between heterogeneous features and potentially introducing redundancy. To address these issues, we propose the SSCFM, which integrates LKA and SSE to achieve adaptive spatial-spectral alignment and dynamic cross-modal interaction, thereby enabling more effective feature fusion and enhancing reconstruction quality.

\section{Methodology}
\label{sec:method}

\begin{figure}[!tb]
    \centering
    \includegraphics[width=1\linewidth]{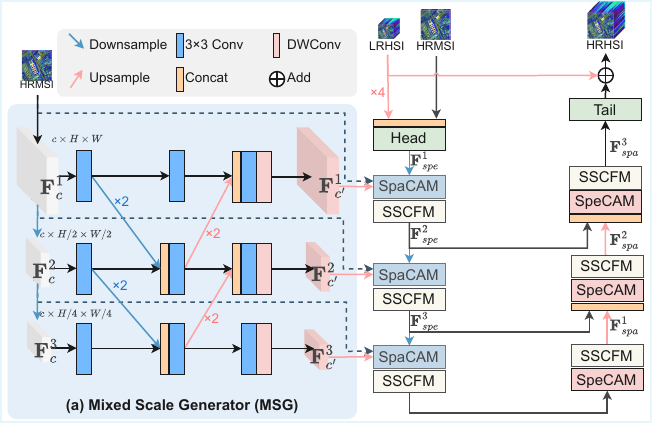}
    \caption{Overall architecture of the proposed CoFusion framework. The framework features a three-level pyramidal pipeline driven by the Multi-Scale Generator (MSG) and employs dual-path coordinate-aware modules for collaborative spatial-spectral reconstruction.}
    \label{fig:cofusion:main}
\end{figure}

\subsection{CoFusion Network Architecture}

As illustrated in Fig.~\ref{fig:cofusion:main}, CoFusion adopts a Multi-Scale Generator (MSG) to construct a three-level pyramidal pipeline, enabling joint optimization of fusion performance under different receptive fields. The framework first performs feature alignment and multi-scale representation learning on the LRHSI and HRMSI to generate multi-scale features $\{\bm{F}_c^i \mid i=1,2,3\}$.

At each scale, a dual-branch architecture is employed for parallel feature extraction. The SpeCAM branch focuses on maintaining spectral fidelity by modeling deep spectral dependencies through the Spectral Coordinate Mixing (SCM) mechanism. Simultaneously, the SpaCAM branch is dedicated to spatial structure reconstruction, capturing multi-scale spatial contexts via Gated Spatial Blocks (GSB) with varying dilation rates.

To enable cross-branch interaction, the Spatial-Spectral Cross-Fusion Module (SSCFM) leverages Large Kernel Attention (LKA) and Spectral Squeeze-and-Excitation (SSE) to perform adaptive feature alignment and fusion. This design balances spatial enhancement and spectral preservation. Finally, the refined multi-scale features are progressively upsampled and aggregated through residual connections, eventually generating the HRHSI via a reconstruction head.

\subsection{MSG: Multi-Scale Generator}

Inspired by HRNet~\cite{HRNet}, the MSG constructs a three-level pyramidal representation to bridge global semantics and local details (Fig.~\ref{fig:cofusion:main}(a)). It generates multi-scale features at scale $l \in \{1,2,3\}$, denoted as local features $\bm{F}_c^l$ and global proxy features $\bm{F}_{c'}^l$.

The MSG maintains parallel feature flows guided by the HRMSI. Specifically, local features are obtained through successive bilinear downsampling ($\downarrow^2$) and encoded via $\bm{E}_c$, while global proxy features are generated by upsampling ($\uparrow^2$) coarser representations and processed by $\bm{E}_{c'}$. Formally:
\begin{equation}
\bm{F}_c^l = \bm{E}_c \bigl( \downarrow^2 (\bm{F}_c^{l-1}) \bigr), \quad
\bm{F}_{c'}^l = \bm{E}_{c'} \bigl( \uparrow^2 (\bm{F}_{c'}^{l+1}) \bigr).
\end{equation}

\subsection{SpaCAM: Spatial Context-Aware Module}

\begin{figure}[!tb]
    \centering
    \includegraphics[width=1\linewidth]{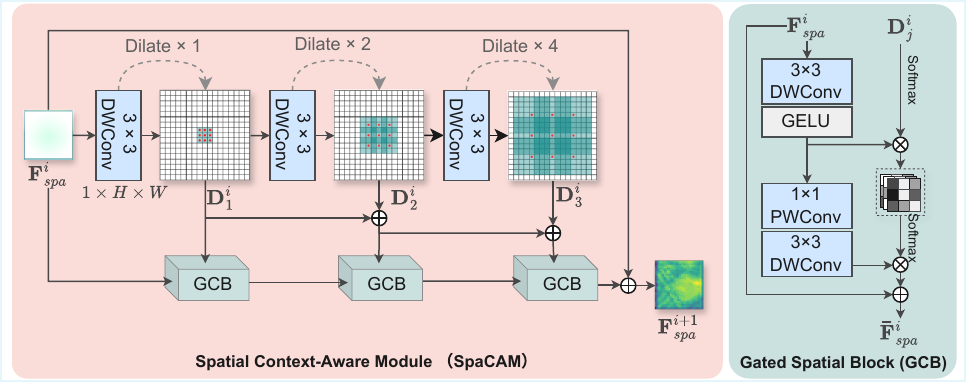}
    \caption{Detailed structure of SpaCAM. It adaptively refines spatial details by leveraging depthwise separable convolutions with multiple dilation rates and the Gated Spatial Block (GSB) mechanism.}
    \label{fig:cofusion:spacam}
\end{figure}

In HSI fusion tasks, the scale variance of ground objects is often extreme, and fine textures such as edges and point targets are highly susceptible to destruction by smoothing operators during spatial enhancement. Traditional fixed-kernel convolutions struggle to balance local detail extraction with global context awareness~\cite{derconv_01, derconv_02} and frequently introduce redundant noise during feature transmission. To address these challenges, we designed SpaCAM to achieve adaptive refinement of spatial information through multi-receptive field feature extraction and a dynamic gating mechanism.

As shown in Fig.~\ref{fig:cofusion:spacam}, this module consists of a series of convolutional layers with different dilation rates and Gated Spatial Blocks (GSB). First, to capture multi-scale spatial contexts, the input feature $\bm{F}_{spa}^{i}$ is processed by depthwise separable convolutions (DWConv) with dilation rates $d \in \{1, 2, 4\}$. By utilizing dilated convolutions to expand the receptive field without increasing parameters, we generate scale-aware feature representations $\mathbf{D}_{j}^{i}$:
{
\small
\begin{equation}
\mathbf{D}_{j}^{i} = \text{DWConv}_{3\times3, d}(\text{MaxPool}(\bm{F}_{spa}^{i})), \quad j \in \{1, 2, 3\}
\end{equation}
}

Subsequently, to suppress background noise and highlight salient object features, these features are fed into corresponding GSBs. Inside each GSB, a Softmax-based gating mechanism is introduced to achieve dynamic attention and feature calibration for important spatial regions:
\begin{equation}
\overline{\bm{F}}_{spa}^{i} = \text{Softmax}(\mathbf{D}_{j}^{i}) \otimes \sigma(\text{DWConv}(\bm{F}_{spa}^{i})) + \bm{F}_{spa}^{i}
\end{equation}
where $\sigma$ denotes the GELU activation function and $\otimes$ denotes element-wise multiplication. This design ensures the network adaptively focuses on salient spatial details in complex geographical scenes, effectively alleviating edge blurring and outputting the enhanced feature $\bm{F}_{spa}^{i+1}$.

\begin{figure}[!tb]
    \centering
    \includegraphics[width=1\linewidth]{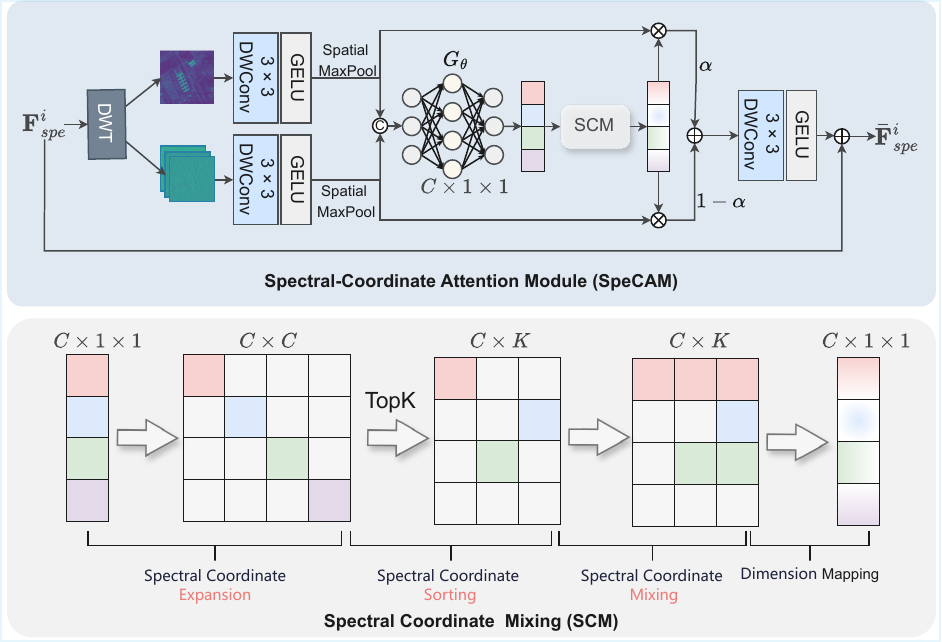}
    \caption{Schematic of SpeCAM and its core Spectral Coordinate Mixing (SCM) operator. This module employs frequency-domain decoupling and a coordinate-based traversal strategy to capture deep spectral correlations.}
    \label{fig:cofusion:specam}
\end{figure}

\begin{algorithm}[!htb]
\caption{SpeCAM Algorithm}
\label{alg:specam}

\SetKwInOut{Input}{Input}
\SetKwInOut{Output}{Output}

\Input{
Spectral feature $\mathbf{F}_{\text{spe}}^{i} \in \mathbb{R}^{H \times W \times C}$;
Mixing weight $\alpha$; Sparsity $K$; Initial state $\mathbf{z}^{0} \in \mathbb{R}^{C}$
}
\Output{Refined feature $\bar{\mathbf{F}}_{\text{spe}}^{i}$}

\BlankLine
\textbf{Stage 1: Frequency Decomposition and Saliency Extraction}

$\{f_{\text{low}}, f_{\text{high}}\} \gets \text{DWT}(\mathbf{F}_{\text{spe}}^{i})$\;

\For{$f \in \{f_{\text{low}}, f_{\text{high}}\}$}{
    $f \gets \text{GELU}(\text{DWConv}(f))$\;
    $\mathbf{v}_f \gets \text{MaxPool}(f)$ \tcp*{$\mathbb{R}^{C \times 1 \times 1}$}
}

\BlankLine
\textbf{Stage 2: Spectral Coordinate Mixing (SCM)}

$\mathbf{z}_{in} \gets G_{\theta}(\text{Concat}(\mathbf{v}_{\text{low}}, \mathbf{v}_{\text{high}}))$\;

$\mathbf{M} \gets \text{Top-K}(\text{Expansion}(\mathbf{z}_{in}), K)$\;

$\mathbf{z} \gets \mathbf{z}^{0}$\;

\For{$k = 1$ \KwTo $K$}{
    $\mathbf{z} \gets \text{Update}(\mathbf{z}, \mathbf{M}(:,k))$\;
}

$\mathbf{a} \gets \text{Map}(\mathbf{z})$\;

\BlankLine
\textbf{Stage 3: Weighted Fusion and Residual Learning}

$\mathbf{F}_{\text{low}} \gets f_{\text{low}} \otimes \mathbf{a} \cdot \alpha$\;

$\mathbf{F}_{\text{high}} \gets f_{\text{high}} \otimes \mathbf{a} \cdot (1 - \alpha)$\;

$\mathbf{F}_{mid} \gets \text{GELU}(\text{DWConv}(\mathbf{F}_{\text{low}} + \mathbf{F}_{\text{high}}))$\;

$\bar{\mathbf{F}}_{\text{spe}}^{i} \gets \mathbf{F}_{mid} + \mathbf{F}_{\text{spe}}^{i}$\;

\Return $\bar{\mathbf{F}}_{\text{spe}}^{i}$\;

\end{algorithm}

Due to the extremely high dimensionality of HSI, traditional attention mechanisms often fail to achieve an ideal balance between computational efficiency and deep spectral correlation modeling. The design of SpeCAM is primarily inspired by two factors: 1) the physical properties of real-world spectral curves are continuous and smooth, and the channel redundancy from discrete sampling contains rich intrinsic structural information; 2) the performance bottleneck of HSI fusion often lies in capturing subtle high-frequency spectral features, which are easily smoothed out by conventional convolutions. Inspired by candidate selection and combination strategies in automatic multi-coordinate update strategies, SpeCAM achieves dynamic reconstruction of critical spectral features through frequency-domain decomposition and coordinate mixing mechanisms, as shown in Fig.~\ref{fig:cofusion:specam}.

To overcome the bottleneck of high-frequency feature modeling, the input feature $\bm{F}_{spe}^{i}$ is first decomposed into a low-frequency component representing the global envelope and a high-frequency component representing detail fluctuations via Discrete Wavelet Transform (DWT)~\cite{DWT}. This frequency-domain decoupling helps the model examine spectral continuity across different scales. Subsequently, the two frequency branches are mapped via DWConv and activation functions, followed by spatial max-pooling to compress dimensions into a vector sequence representing channel saliency.

The SCM mechanism is the core of SpeCAM. Within SCM, feature processing involves three stages: 1) \textit{Spectral Coordinate Expansion}: by modeling global correlations among all channels, a $C \times C$ interaction space is constructed to mine continuous spectral features; 2) \textit{Spectral Coordinate Ranking}: a Top-K operator is used to filter the top $K$ key coordinates, generating a $C \times K$ spectral coordinate matrix to aggregate salient innovative information; 3) \textit{Spectral Coordinate Mixing}: coordinate traversal and innovation are performed on the matrix from left to right, as detailed in Algorithm~\ref{alg:specam}. The update process for the $c$-th channel at the $k$-th candidate coordinate follows these criteria:

\begin{equation}
\bar{\mathbf{z}}_{c}^{t} = 
\begin{cases} 
\mathbf{z}_{c}^{t}, & \mathbf{z}_{c}^{t} \neq 0 \\
\mathbf{z}_{c}^{t-1}, & \mathbf{z}_{c}^{t} = 0
\end{cases}
\end{equation}

where $c \in \{1, \dots, C\}$ and $k \in \{1, \dots, K\}$. This formula represents the dynamic innovation logic of spectral coordinates: when traversing the $C \times K$ matrix, if a non-zero innovation value $\mathbf{z}_{c}^{t}$ is detected, the system accepts this new value and updates the state, thereby capturing local mutations (high-frequency features) in the spectral curve. If the value is zero, the previous state $\mathbf{z}_{c}^{t-1}$ is retained to maintain physical spectral continuity. This traversal-based filtering mechanism ensures the model adaptively extracts the most relevant spectral features from the sparse matrix. Finally, complementary fusion of high and low-frequency branches is achieved via learnable coefficients $\alpha$ and $1-\alpha$, and the refined spectral feature $\bar{\bm{F}}_{spe}^{i}$ is output via residual connection.

\subsection{SSCFM: Spatial-Spectral Cross-Fusion Module}

\begin{figure}[!tb]
    \centering
    \includegraphics[width=1\linewidth]{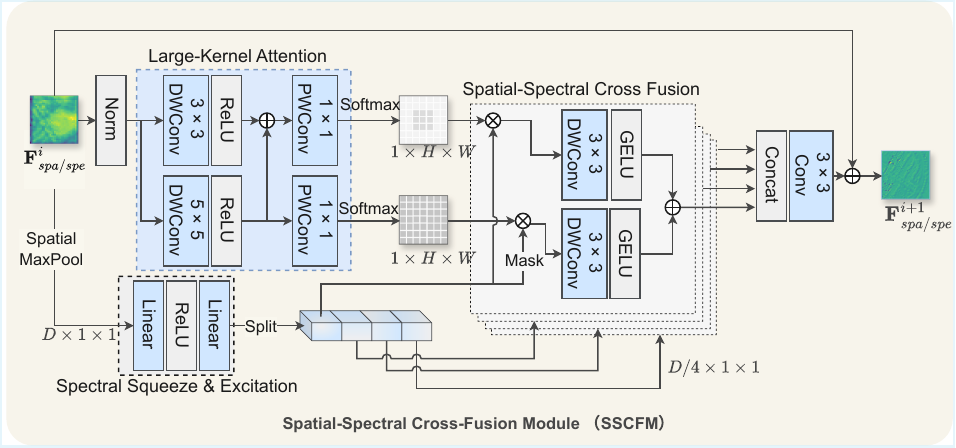}
    \caption{Architecture of SSCFM. It utilizes spatial and spectral branches to achieve fine-grained inter-modal interaction and feature alignment.}
    \label{fig:cofusion:sscfm}
\end{figure}

After obtaining enhanced spatial features and refined spectral features, the key to improving fusion quality lies in achieving deep decoupling and collaborative mapping between the two. We designed the SSCFM, whose core motivation stems from the complementarity and redundancy of cross-modal features in representing geographical entities. Specifically, while spatial features contain rich textural details, they are susceptible to spectral distortion; conversely, spectral features possess physical attribute consistency but lack sufficient spatial resolution. Thus, SSCFM aims to eliminate representational redundancy between modalities and compensate for missing critical details through a cross-modal interaction mechanism. As shown in Fig.~\ref{fig:cofusion:sscfm}, this module is jointly driven by Large Kernel Attention (LKA) and Spectral Squeeze-and-Excitation (SSE)~\cite{SENet} branches to complete the final spatial-spectral cross-fusion.

Initially, input features $\bm{F}_{spa/spe}^{i}$ undergo Layer Normalization (Norm) before entering the LKA branch, where multi-granularity spatial context correlations are captured via parallel $3 \times 3$ and $5 \times 5$ DWConvs. Subsequently, a Softmax operator is used to generate spatial attention masks at different scales to guide precise spatial information selection. Simultaneously, $\bm{F}_{spa/spe}^{i}$ is compressed into a $D \times 1 \times 1$ channel vector via spatial max-pooling, followed by linear layers and ReLU activation to achieve spectral reconstruction and information excitation, generating the spectral-guided feature vector $\bm{V}$.
\begin{table*}[htb]
\centering
\caption{Quantitative comparison of various methods on the Chikusei, CAVE, and PaviaU datasets across $\times 2$, $\times 4$, and $\times 8$ scaling factors. \textbf{Bold} indicates the best performance, while \underline{underline} denotes the second-best.}
\label{tab:main}
\resizebox{\textwidth}{!}{
\begin{tabular}{l|c|cccc|cccc|cccc}
\toprule
\multirow{2}{*}{Method} & \multirow{2}{*}{Scale}   
& \multicolumn{4}{c|}{Chikusei} 
& \multicolumn{4}{c|}{CAVE} 
& \multicolumn{4}{c}{PaviaU} \\
& & PSNR$\uparrow$ & SSIM$\uparrow$ & SAM$\downarrow$ & ERGAS$\downarrow$   
& PSNR$\uparrow$ & SSIM$\uparrow$ & SAM$\downarrow$ & ERGAS$\downarrow$   
& PSNR$\uparrow$ & SSIM$\uparrow$ & SAM$\downarrow$ & ERGAS$\downarrow$ \\ 
\midrule

HSRnet~\cite{HSRnet} & $\times 2$
& 51.5324 & 0.9951 & 2.0982 & 1.5893
& 50.5410 & 0.9926 & 2.4608 & 2.0012
& 39.4947 & 0.9662 & 3.9012 & 2.4531 \\

Fusformer~\cite{Fusformer} & $\times 2$
& 52.3104 & 0.9954 & 2.0596 & 1.5698
& 51.3352 & 0.9935 & 2.3294 & 1.9692
& 40.3442 & 0.9703 & 3.5187 & 2.2514 \\

PSRT~\cite{PSRT} & $\times 2$
& 52.3821 & 0.9957 & 2.0091 & 1.4692
& 51.4128 & 0.9941 & 2.1795 & 1.8891
& 40.4956 & 0.9766 & 2.3996 & 2.0812 \\

U2Net~\cite{U2Net} & $\times 2$
& \underline{52.4657} & 0.9961 & 2.0003 & 1.4098
& 51.4816 & 0.9948 & 2.1297 & 1.8496
& \underline{40.7347} & \underline{0.9789} & \underline{2.3092} & \underline{1.5398} \\

3DT-Net~\cite{3DT-Net} & $\times 2$
& 52.4155 & \underline{0.9974} & \underline{1.9498} & 1.4195
& \underline{51.5043} & \underline{0.9958} & \underline{2.0496} & \underline{1.8294}
& 39.7224 & 0.9784 & 2.4692 & 1.7896 \\

BUGPan~\cite{BUGPan} & $\times 2$
& 51.6205 & 0.9957 & 2.1194 & 1.5902
& 50.5846 & 0.9933 & 2.4992 & 2.0418
& 39.5856 & 0.9724 & 3.2294 & 2.2715 \\

FMPM-DNet~\cite{FMPM-DNet} & $\times 2$
& 51.9826 & 0.9962 & 2.0497 & 1.4996
& 50.9621 & 0.9943 & 2.2693 & 1.9397
& 40.2562 & 0.9767 & 2.9594 & 2.1312 \\

SMGU-Net~\cite{SMGU-Net} & $\times 2$
& 51.2751 & 0.9971 & 2.0195 & \underline{1.3897}
& 51.5052 & 0.9956 & 2.0792 & 1.8194
& 37.8881 & 0.9753 & 3.0192 & 2.0496 \\

\rowcolor{gray!15}
CoFusion & $\times 2$
& \textbf{53.0875} & \textbf{0.9981} & \textbf{1.8756} & \textbf{1.2536}
& \textbf{51.5247} & \textbf{0.9972} & \textbf{2.0238} & \textbf{1.7421}
& \textbf{41.3824} & \textbf{0.9872} & \textbf{1.8341} & \textbf{1.2268} \\

\midrule

HSRnet~\cite{HSRnet} & $\times 4$
& 47.8521 & 0.9884 & 2.6137 & 2.0845
& 46.8219 & 0.9845 & 3.1328 & 2.5831
& 35.7642 & 0.9517 & 4.0413 & 3.4476 \\

Fusformer~\cite{Fusformer} & $\times 4$
& 48.6413 & 0.9892 & 2.5481 & 2.0429
& 47.6254 & 0.9852 & 3.0047 & 2.4716
& 36.2917 & 0.9531 & 4.1105 & 3.1428 \\

PSRT~\cite{PSRT} & $\times 4$
& 49.0346 & 0.9905 & 2.3518 & 1.9247
& 47.9921 & 0.9863 & 2.8354 & 2.2913
& 36.7725 & 0.9628 & 3.5932 & 2.6541 \\

U2Net~\cite{U2Net} & $\times 4$
& 49.0028 & 0.9913 & 2.3245 & 1.8617
& 47.9315 & 0.9872 & 2.7719 & 2.1634
& 37.2641 & 0.9672 & 3.2358 & 2.4015 \\

3DT-Net~\cite{3DT-Net} & $\times 4$
& 46.6914 & 0.9941 & 2.5832 & 2.3218
& 45.6013 & 0.9904 & 3.1947 & 2.8312
& 36.6821 & 0.9725 & \underline{3.0416} & 2.5327 \\

BUGPan~\cite{BUGPan} & $\times 4$
& 48.3217 & 0.9903 & 2.4756 & 2.1432
& 47.2612 & 0.9875 & 2.9731 & 2.5518
& 37.0654 & 0.9612 & 3.8427 & 2.9613 \\

FMPM-DNet~\cite{FMPM-DNet} & $\times 4$
& 48.7125 & 0.9914 & 2.3918 & 1.9925
& 47.7058 & 0.9882 & 2.8841 & 2.3521
& \underline{37.7113} & 0.9653 & 3.4618 & 2.7052 \\

SMGU-Net~\cite{SMGU-Net} & $\times 4$
& \underline{49.8316} & \underline{0.9962} & \underline{2.2531} & \underline{1.7428}
& \underline{48.8157} & \underline{0.9934} & \underline{2.7215} & \underline{2.0531}
& 37.4258 & \underline{0.9731} & 3.2045 & \underline{2.1643} \\

\rowcolor{gray!15}
CoFusion & $\times 4$
& \textbf{50.6742} & \textbf{0.9971} & \textbf{2.1494} & \textbf{1.7252}
& \textbf{49.1382} & \textbf{0.9948} & \textbf{2.6045} & \textbf{2.0087}
& \textbf{38.3195} & \textbf{0.9824} & \textbf{2.5637} & \textbf{1.9482} \\

\midrule

HSRnet~\cite{HSRnet} & $\times 8$
& 46.2231 & 0.9882 & 2.8054 & 2.5317
& 45.2015 & 0.9831 & 3.3621 & 3.0854
& 28.6942 & 0.8921 & 6.4913 & 5.4128 \\ 

Fusformer~\cite{Fusformer} & $\times 8$
& 46.3947 & 0.9893 & 2.7415 & 2.4358
& 45.3452 & 0.9853 & 3.2847 & 2.9631
& 28.9813 & 0.8982 & 6.4152 & 5.3947 \\ 

PSRT~\cite{PSRT} & $\times 8$
& 47.1428 & 0.9921 & 2.6951 & 2.2134
& 46.1157 & 0.9872 & 3.1658 & 2.7215
& 30.2642 & 0.9161 & 6.0825 & 5.3218 \\ 

U2Net~\cite{U2Net} & $\times 8$
& 47.3945 & 0.9945 & 2.5518 & 2.1132
& 46.3618 & 0.9892 & 3.0354 & 2.6317
& \underline{30.4251} & 0.9184 & 5.9328 & 5.0715 \\ 

3DT-Net~\cite{3DT-Net} & $\times 8$
& 46.1816 & 0.9895 & 2.9231 & 2.5014
& 45.3147 & 0.9856 & 3.5042 & 3.1358
& 29.4851 & 0.9042 & 6.1547 & 5.3321 \\

BUGPan~\cite{BUGPan} & $\times 8$
& 46.8613 & 0.9897 & 2.9654 & 2.5081
& 45.7819 & 0.9842 & 3.4215 & 3.0162
& 29.4817 & 0.9128 & 6.2631 & 5.4182 \\

FMPM-DNet~\cite{FMPM-DNet} & $\times 8$
& 47.2851 & 0.9908 & 2.8137 & 2.2945
& 46.2346 & 0.9861 & 3.2158 & 2.8213
& 30.1812 & 0.9175 & \underline{5.8143} & 5.0429 \\

SMGU-Net~\cite{SMGU-Net} & $\times 8$
& \underline{48.1512} & \underline{0.9954} & \underline{2.4831} & \underline{1.9315}
& \underline{47.1354} & \underline{0.9911} & \underline{2.9054} & \underline{2.4132}
& 28.2813 & \underline{0.9382} & 5.8817 & \underline{4.8516} \\ 

\rowcolor{gray!15}
CoFusion & $\times 8$
& \textbf{48.9371} & \textbf{0.9963} & \textbf{2.2156} & \textbf{1.8124}
& \textbf{47.4931} & \textbf{0.9934} & \textbf{2.8276} & \textbf{2.3552}
& \textbf{30.7428} & \textbf{0.9481} & \textbf{5.7142} & \textbf{4.5213} \\

\bottomrule
\end{tabular}
}
\end{table*}

To maintain a lightweight architecture while enhancing the granularity of feature interaction, we adopt a channel-wise split strategy. Specifically, vector $\bm{V}$ is uniformly divided into four segments (each with dimension $D/4 \times 1 \times 1$), which are then fed into the spatial-spectral cross-fusion operator. This design not only reduces the computational complexity of cross-modal point-wise multiplication but also enables the model to explore local correlations within smaller feature subspaces. Additionally, considering potential noise interference during sensor imaging, a 10\% random Dropout mechanism (implemented via masking) is introduced during interaction with large-scale spatial attention maps. By simulating feature absence, the model is forced to learn more robust fusion logic. Finally, cross-feature flows are re-aggregated via concatenation (Concat) after convolutional calculation, and then output as the final residual map through a $3 \times 3$ convolutional layer to achieve deep refinement of $\bm{F}_{spa/spe}^{i}$.

\subsection{Loss Function}

To guide the network in learning fused features with high spatial resolution and high spectral fidelity, CoFusion employs a joint loss function for end-to-end training. The total loss consists of a weighted combination of $\mathcal{L}_1$ loss (pixel-level error) and Structural Similarity (SSIM) loss, aiming to constrain the fusion results to approximate the Ground Truth (GT) in both numerical intensity and structural texture.

\paragraph{$\mathcal{L}_1$ Loss.} We calculate the pixel-wise Mean Absolute Error (MAE) between the final fused image $\bar{\mathbf{Z}}$ and the reference GT image $\mathbf{Z}$ to ensure numerical fidelity:
{\small
\begin{equation}
\mathcal{L}_{1} = \frac{1}{C \times H \times W} \sum_{c=1}^{C} \sum_{h=1}^{H} \sum_{w=1}^{W} | \bar{\mathbf{Z}}(c,h,w) - \mathbf{Z}(c,h,w) |
\end{equation}
}
where $\bar{\mathbf{Z}}$ is the final result generated by the proposed network based on input LRHSI $\mathbf{X}$ and HRMSI $\mathbf{Y}$.

\paragraph{SSIM Loss.} SSIM measures the difference between $\bar{\mathbf{Z}}$ and $\mathbf{Z}$ from three dimensions: luminance, contrast, and structure. For the $c$-th band, SSIM is defined as:

{
\small
\begin{equation}
\text{SSIM}(\mathbf{x}_c, \mathbf{y}_c) = \frac{(2\mu_{\mathbf{x}_c}\mu_{\mathbf{y}_c} + C_1)(2\sigma_{\mathbf{x}_c\mathbf{y}_c} + C_2)}{(\mu_{\mathbf{x}_c}^2 + \mu_{\mathbf{y}_c}^2 + C_1)(\sigma_{\mathbf{x}_c}^2 + \sigma_{\mathbf{y}_c}^2 + C_2)}
\end{equation}
where $\mathbf{x}_c$ and $\mathbf{y}_c$ denote the local window patches of the fused and reference images in the $c$-th band, respectively. The SSIM loss is constructed by averaging the SSIM values across all bands:
\begin{equation}
\mathcal{L}_{ssim} = 1 - \frac{1}{C} \sum_{c=1}^{C} \text{SSIM}(\bar{\mathbf{Z}}_{c}, \mathbf{Z}_{c})
\end{equation}
}

\paragraph{Total Loss Function.} Network parameters are optimized in an end-to-end manner. The total loss function is integrated by the weighted sum of the two aforementioned terms:
\begin{equation}
\mathcal{L}_{total} = \mathcal{L}_{1} + \lambda \cdot \mathcal{L}_{ssim}
\end{equation}
where the trade-off coefficient $\lambda$ is empirically fixed at $0.1$. This joint loss function balances pixel-level accuracy and spatial structural integrity, ensuring the fused image retains the spectral information of $\mathbf{X}$ while integrating high-frequency spatial details from $\mathbf{Y}$.

\section{Experimental Results and Analysis}
\label{sec:experiments}

\begin{figure*}[htb]
    \centering
    \includegraphics[width=1\linewidth]{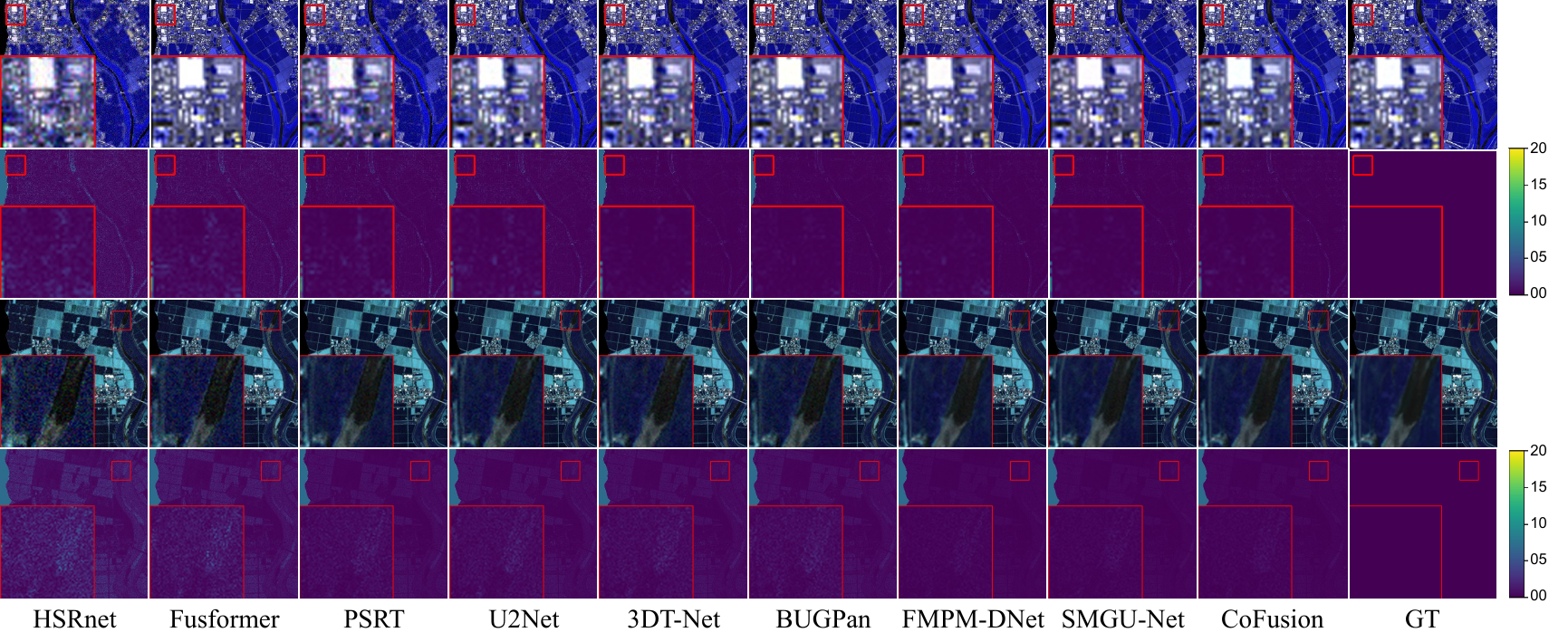}
    \caption{Pseudo-color images with spectral bands 67-15-13 (first patch) and 64-58-16 (third patch) as RGB of $\times$4 fusion results, and corresponding SAM error maps produced by all methods on the Chikusei dataset.}
    \label{fig:cofusion:chikusei}
\end{figure*}

\subsection{Experimental Settings}
\label{exp}

\begin{figure*}[tbh]
    \centering
    \includegraphics[width=1\linewidth]{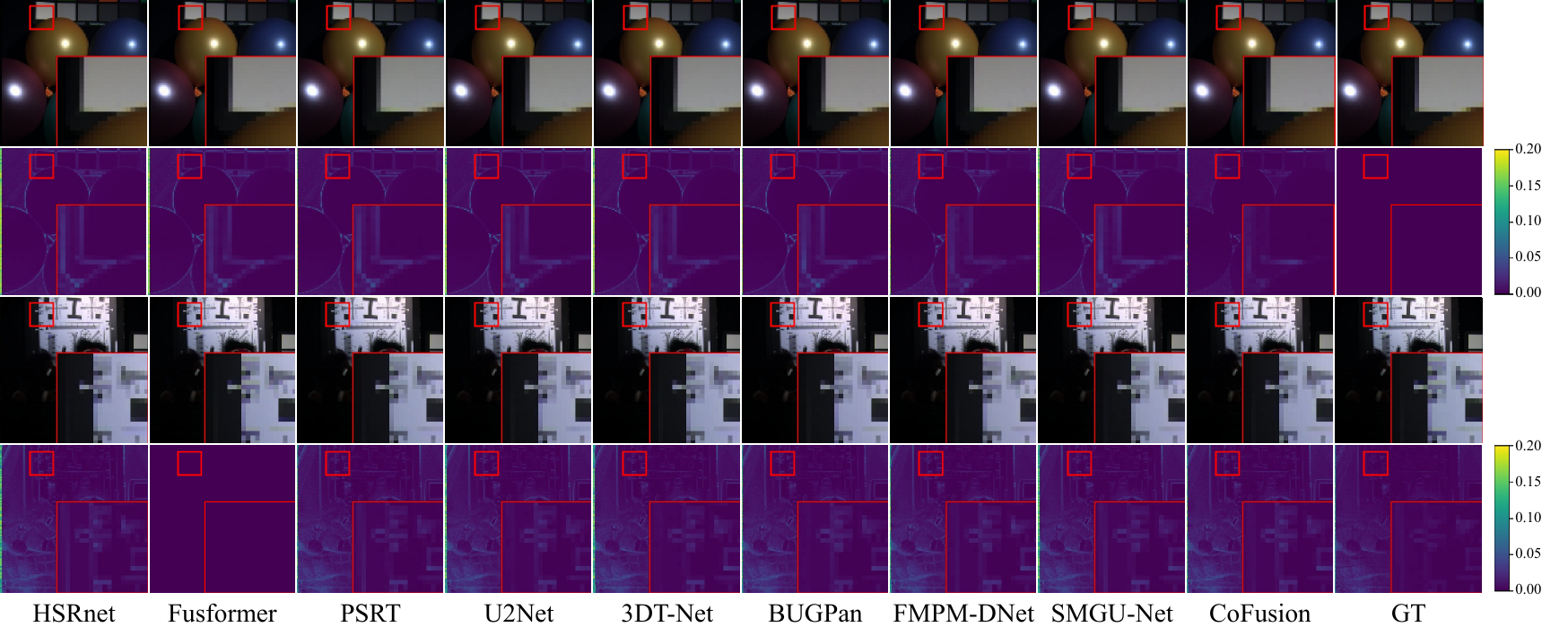}
  \caption{Pseudo-color images with spectral bands 20-10-3 (\textit{balloons}) and 6-25-11 (\textit{Chart and Stuffed Toy}) as RGB of $\times$4 fusion results, and corresponding SAM error maps produced by all methods on the CAVE dataset.}
    \label{fig:cofusion:cave}
\end{figure*}

\paragraph{Datasets.}
To comprehensively validate the performance of the proposed method in spatial reconstruction and spectral fidelity, we conducted extensive experiments on three representative benchmark datasets: Chikusei~\cite{Chikusei}, PaviaU, and CAVE. These datasets cover a variety of application scenarios ranging from large-scale outdoor remote sensing monitoring to controlled indoor imaging, providing robust support for evaluating the model's cross-domain generalization capability. Specifically, Chikusei is a large-scale aerial hyperspectral dataset collected from a mixed agricultural-urban area, containing $2517 \times 2335$ pixels and 128 spectral bands (363--1018 nm); its complex spatial structures and subtle land-cover transitions pose a severe challenge to the ability to recover fine details under significant spatial degradation. The PaviaU dataset, acquired by the ROSIS-03 sensor, consists of $610 \times 340$ pixels and 103 spectral bands. Its abundant urban elements, such as buildings, roads, and vegetation, introduce significant spatial discontinuities and spectral variability, making it an ideal benchmark for evaluating spatial-spectral consistency in structured scenes. In contrast, the CAVE dataset, containing 32 indoor scenes, provides nearly noise-free ground truth (GT), allowing for precise measurement of the absolute accuracy of spectral reconstruction, with an evaluation focus on the model's ability to preserve intrinsic spectral characteristics. Regarding data simulation, we followed Wald's protocol, simulating HRMSI via spectral response functions and constructing LRHSI using Gaussian blurring followed by $4\times$ spatial downsampling.

\paragraph{Comparison Methods and Implementation Details.}
We conducted an in-depth comparison of CoFusion with eight representative state-of-the-art methods. These algorithms encompass various mainstream architectural paradigms, including CNN-based models (HSRnet~\cite{HSRnet}), Transformer-based architectures (Fusformer~\cite{Fusformer}, PSRT~\cite{PSRT}), U-Net variants (U2Net~\cite{U2Net}), and hybrid design schemes (BUGPan~\cite{BUGPan}, FMPM-DNet~\cite{FMPM-DNet}, SMGU-Net~\cite{SMGU-Net}). To ensure experimental fairness, all competing algorithms followed the default hyperparameter configurations provided by their original authors. These methods represent different design philosophies in current spatial-spectral modeling, providing a comprehensive benchmark for evaluating the effectiveness of explicit cross-modal collaboration.

In terms of evaluation metrics, we employed PSNR and SSIM to measure spatial reconstruction quality, while utilizing SAM and ERGAS to assess spectral fidelity. This dual evaluation mechanism is crucial for MHIF tasks, as hyperspectral image fusion inherently involves a trade-off between spatial sharpness and spectral consistency. CoFusion is implemented based on the PyTorch framework and trained end-to-end using the AdamW optimizer. The model is configured with a compact hidden dimension ($D=64$), emphasizing the efficiency of feature interaction rather than increasing capacity through simple parameter stacking. This design philosophy reflects our core idea of improving feature representation quality through a structured spatial-spectral collaborative mechanism.

\subsection{Quantitative Results Analysis}
We performed a comprehensive evaluation of CoFusion across all benchmark datasets for $\times2$, $\times4$, and $\times8$ scaling factors. As shown in Table~\ref{tab:main}, CoFusion consistently achieves optimal performance across various datasets and scaling factors. This strongly demonstrates that explicit modeling of spatial-spectral collaboration can effectively address the inherent ambiguity in hyperspectral fusion, successfully achieving a balance between the competing objectives of spatial detail recovery and spectral fidelity preservation.

\begin{figure*}[htb]
    \centering
    \includegraphics[width=1\linewidth]{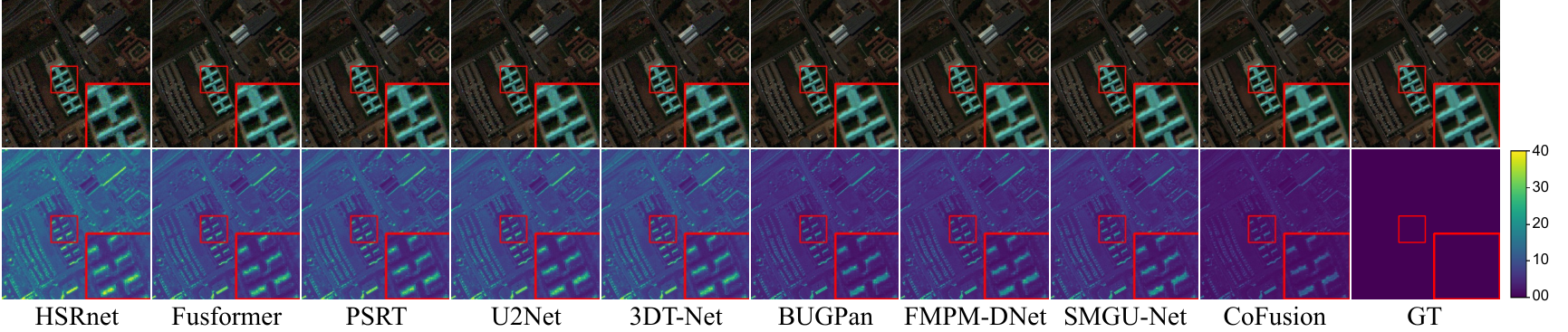}
    \caption{Pseudo-color images with spectral bands 57-27-9 as RGB of $\times$4 fusion results, and corresponding SAM error maps produced by all methods on the PaviaU dataset.}
    \label{fig:cofusion:paviau}
\end{figure*}

\paragraph{Chikusei Dataset Analysis.} On the large-scale remote sensing Chikusei dataset, CoFusion demonstrates excellent robustness under the $\times8$ super-resolution setting. Under this degradation condition, severe information loss increases the difficulty of spatial reconstruction. The improvement in experimental results indicates that SpaCAM helps reconstruct structural priors, while SSCFM enhances the model's reconstruction stability and structural integrity in harsh degradation environments by integrating multi-scale feature cues.

\paragraph{PaviaU Dataset Analysis.} On the PaviaU dataset, CoFusion achieves steady gains in PSNR and SSIM metrics, reflecting its ability to maintain structural consistency in complex urban scenes. Furthermore, the improvement in spectral metrics suggests that spatial enhancement does not lead to significant spectral distortion, indicating that the proposed collaborative mechanism achieves a balance between spatial sharpening and spectral preservation to a certain extent.

\paragraph{CAVE Dataset Analysis.} On the CAVE dataset, CoFusion maintains superior spectral fidelity. Given that the spatial variation in this dataset is relatively smooth, it is primarily used to evaluate the model's spectral representation capability. The experimental performance demonstrates that SpeCAM effectively captures long-range spectral dependencies and maintains spectral continuity during the feature mapping process.

\paragraph{Performance on Real-world Scenarios.}
To evaluate the practical applicability and generalization capability of CoFusion, we conducted experiments on real-world datasets where the ground truth is unavailable. In this unsupervised setting, we employ the Quality with No Reference (QNR) index, which consists of the spectral distortion index $D_\lambda$ and the spatial distortion index $D_s$. As summarized in Table~\ref{tab:results_x4}, CoFusion consistently achieves the lowest distortion values ($D_\lambda, D_s$) and the highest $QNR$ scores across the Chikusei, CAVE, and PaviaU datasets. Specifically, compared with Transformer-based methods like PSRT and hybrid models such as SMGU-Net, our method exhibits superior performance in preserving spectral consistency while significantly sharpening spatial structures. These results demonstrate that the proposed spatial-spectral collaborative modeling effectively captures the underlying physical characteristics of the sensors and maintains a robust fusion logic, even when explicit supervision is absent.

\begin{table*}[htbp]
\centering
\caption{Quantitative evaluation on real-world benchmarks (Chikusei, CAVE, and PaviaU) under $\times 4$ scaling factors. The \textbf{bold} and \underline{underlined} results represent the best and second-best performance, respectively.}
\label{tab:results_x4}
\setlength\tabcolsep{14pt} 

\resizebox{\textwidth}{!}{
\begin{tabular}{l|ccc|ccc|ccc}
\toprule 
\multirow{2}{*}{Method} 
& \multicolumn{3}{c|}{Chikusei} 
& \multicolumn{3}{c|}{CAVE} 
& \multicolumn{3}{c}{PaviaU} \\
& $D_\lambda \downarrow$ & $D_s \downarrow$ & $QNR \uparrow$ 
& $D_\lambda \downarrow$ & $D_s \downarrow$ & $QNR \uparrow$ 
& $D_\lambda \downarrow$ & $D_s \downarrow$ & $QNR \uparrow$ \\ 
\midrule

HSRnet~\cite{HSRnet}        
& 0.0842 & 0.0915 & 0.8321 
& 0.0921 & 0.1045 & 0.8142 
& 0.1124 & 0.1256 & 0.7785 \\

Fusformer~\cite{Fusformer}
& 0.0795 & 0.0882 & 0.8456 
& 0.0884 & 0.0998 & 0.8265 
& 0.1085 & 0.1198 & 0.7892 \\

PSRT~\cite{PSRT}          
& 0.0768 & 0.0854 & 0.8532 
& 0.0852 & 0.0965 & 0.8348 
& 0.1042 & 0.1145 & 0.7965 \\

U2Net~\cite{U2Net}        
& \underline{0.0685} & 0.0792 & \underline{0.8698} 
& \underline{0.0765} & 0.0885 & \underline{0.8512} 
& \underline{0.0945} & 0.1052 & \underline{0.8124} \\

3DT-Net~\cite{3DT-Net}    
& 0.0712 & \underline{0.0765} & 0.8645 
& 0.0798 & 0.0862 & 0.8465 
& 0.0982 & \underline{0.1025} & 0.8056 \\

BUGPan~\cite{BUGPan}      
& 0.0815 & 0.0895 & 0.8398 
& 0.0905 & 0.1012 & 0.8198 
& 0.1105 & 0.1215 & 0.7825 \\

FMPM-DNet~\cite{FMPM-DNet}
& 0.0745 & 0.0825 & 0.8585 
& 0.0825 & 0.0935 & 0.8412 
& 0.1015 & 0.1112 & 0.8015 \\

SMGU-Net~\cite{SMGU-Net}  
& 0.0695 & 0.0785 & 0.8652 
& 0.0782 & \underline{0.0892} & 0.8485 
& 0.0965 & 0.1065 & 0.8095 \\

\rowcolor{gray!15} 
CoFusion (Ours)          
& \textbf{0.0638} & \textbf{0.0712} & \textbf{0.8768} 
& \textbf{0.0715} & \textbf{0.0810} & \textbf{0.8604} 
& \textbf{0.0882} & \textbf{0.0956} & \textbf{0.8223} \\

\bottomrule
\end{tabular}
}
\end{table*}

In summary, CoFusion demonstrates a remarkable capacity to simultaneously improve spectral fidelity and spatial resolution. In contrast to traditional model-based approaches where these two objectives often pose a competitive trade-off, CoFusion achieves synchronized optimization through its collaborative architectural design. The consistent gains across both supervised and unsupervised metrics suggest that explicit spatial-spectral interaction modeling is a highly effective paradigm for mitigating the fundamental dilemmas in hyperspectral image fusion.

\subsection{Visual Comparison and Spectral Curve Analysis}

\begin{figure*}
    \centering
    \includegraphics[width=1\linewidth]{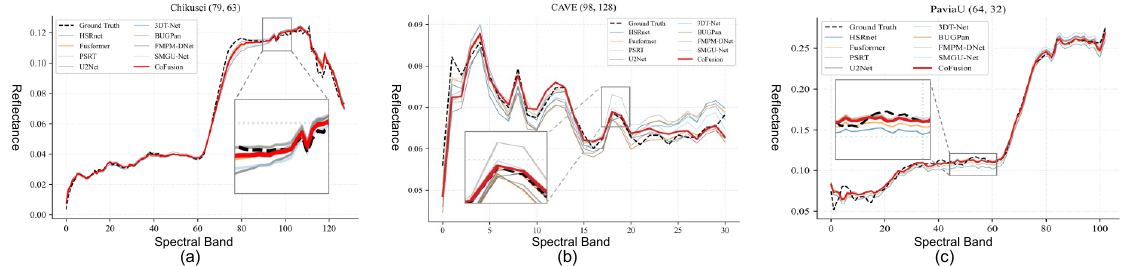}
    \caption{Spectral reconstruction error analysis for three datasets. The subplots from left to right display pixel-level spectral response curves at specific locations: (a) region (79, 63) of the Chikusei dataset; (b) pixel (98, 128) in the ``\textit{Chart and Stuffed Toy}'' scene of the CAVE dataset; and (c) pixel (64, 72) of the PaviaU dataset. The results show that the proposed method (red line) maintains minimal deviation from the ground truth (GT) across the entire spectral range, validating its superior spectral fidelity.}
    \label{fig:cofusion_spectral_1x3}
\end{figure*}

Fig.~\ref{fig:cofusion:chikusei} to Fig.~\ref{fig:cofusion:cave} present the qualitative comparison results for each method. From the spatial perspective, CoFusion reconstructs sharper edges and richer textural details, exhibiting a clear visual advantage especially in high-frequency regions such as object boundaries. In contrast, CNN-based methods, limited by local receptive fields, often produce over-smoothed results, while some attention-based methods are prone to introducing spatial inconsistency artifacts when handling complex scene transitions.

Analyzing from the spectral dimension, the SAM error maps reveal that competing methods generally suffer from significant error accumulation in boundary regions, reflecting their insufficient spatial-spectral alignment capabilities. In sharp contrast, CoFusion produces a more uniform error distribution, indicating that its spatial detail enhancement process strictly adheres to spectral consistency constraints. The spectral reflectance curves in Fig.~\ref{fig:cofusion_spectral_1x3} further validate this conclusion: CoFusion maintains a high degree of fit with the ground truth across the entire band range, proving that fine-grained spatial improvement is not achieved at the expense of spectral integrity.

\subsection{Model Efficiency Analysis}

\begin{figure}[tbh]
    \centering
    \includegraphics[width=1\linewidth]{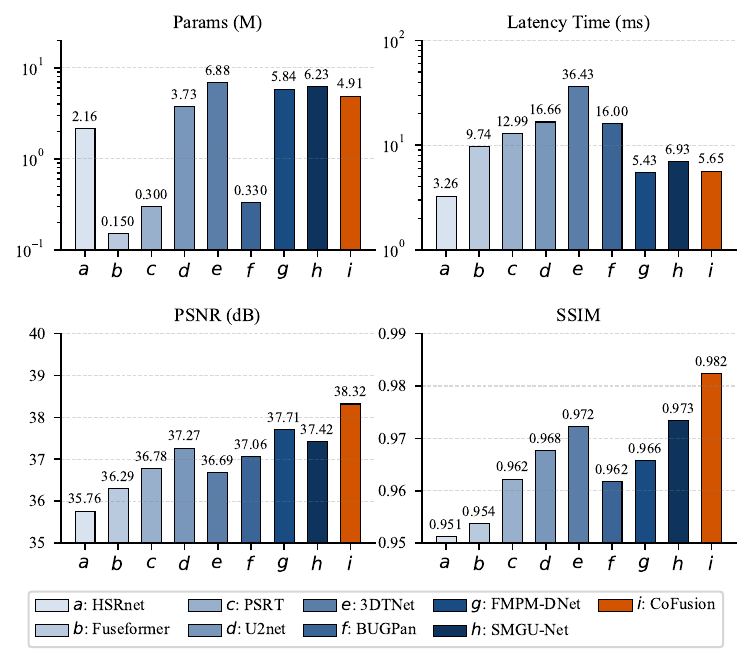}
     \caption{Model efficiency comparison for $\times 4$ fusion on the PaviaU dataset. The bar chart compares the parameter count (in millions, M) and inference latency (in milliseconds, ms, measured on an Intel Xeon CPU). Note that both axes utilize a logarithmic scale to accommodate the wide range of values.}
    \label{fig:cofusion:eff}
\end{figure}

Beyond reconstruction accuracy, we further evaluated the computational efficiency of the model, which is of great significance for practical deployment. Fig.~\ref{fig:cofusion:eff} illustrates the comparison results of CoFusion with other SOTA methods in terms of parameter count (Params) and inference latency (Latency).

The experimental results indicate that CoFusion achieves an excellent balance between performance and efficiency. Regarding inference speed, CoFusion's inference latency is only 5.65 ms, which is significantly better than models such as 3DTNet (36.43 ms) and U2Net (16.66 ms). In terms of model scale, CoFusion controls the parameter count at 4.91 M; while maintaining a smaller model scale, its PSNR (38.32 dB) and SSIM (0.982) are significantly superior to FMPM-DNet (5.84 M) and SMGU-Net (6.23 M), which have larger parameter counts.

This efficiency advantage is primarily attributed to two core design principles: 1) adopting lightweight depthwise separable convolutions for spatial encoding, which effectively reduces computational complexity; 2) avoiding redundant cross-modal repeated calculations through a structured feature routing mechanism. In summary, the performance improvement of CoFusion stems primarily from more effective feature utilization rather than simple parameter stacking, a characteristic that holds high practical value for resource-constrained real-time hyperspectral imaging applications.

\subsection{Ablation Study}

\begin{table}[tbp]
\centering
\caption{Ablation study of CoFusion modules on the PaviaU dataset for $\times4$ fusion task. The \textbf{bold} and \underline{underlined} results represent the best and second-best performance, respectively.}
\label{tab:ablation_study}
\resizebox{\columnwidth}{!}{
\begin{tabular}{l|cc|cc}
\toprule
Variants & PSNR (dB) $\uparrow$ & SSIM $\uparrow$ & Params (M) $\downarrow$ & Latency (ms) $\downarrow$ \\
\midrule
w/o SpaCAM  & 37.1245 & 0.9612 & \textbf{3.58} & \textbf{4.12} \\
w/o SpeCAM  & 37.5862 & 0.9694 & \underline{4.15} & \underline{4.85} \\
w/o SSCFM   & \underline{37.9518} & \underline{0.9752} & 4.62 & 5.28 \\
\rowcolor{gray!15}
Full Model  & \textbf{38.3195} & \textbf{0.9824} & 4.91 & 5.65 \\
\bottomrule
\end{tabular}
}
\end{table}

To validate the design motivation of each core module and its contribution to the final performance, we conducted an ablation study. Specifically, we constructed different variant models by progressively replacing the core components with standard residual blocks. Table~\ref{tab:ablation_study} presents the quantitative results for each variant on the PaviaU dataset, with specific analysis as follows:

The ablation results highlight a clear functional deconstruction of CoFusion: removing SpaCAM causes the most severe PSNR drop (from 38.3195 dB to 37.1245 dB), confirming its fundamental role in recovering high-frequency details and maintaining image topology. Similarly, the ablation of SpeCAM leads to a 0.73 dB decline and obvious spectral distortion, underscoring its criticality in preserving physical properties and cross-band consistency. Furthermore, despite its minimal computational overhead, the removal of SSCFM triggers a systematic performance decline, revealing its indispensable role in multi-scale alignment and interaction stabilization. Ultimately, the organic integration of these specialized modules constitutes CoFusion’s core advantage in balancing high-precision reconstruction with computational efficiency.

\section{Conclusion}
\label{sec:conclusion}

In this paper, we proposed CoFusion: a spatial-spectral collaborative framework for Multispectral and Hyperspectral Image Fusion (MHIF). By explicitly modeling cross-scale interactions and spatial-spectral coupling relationships, the proposed method effectively alleviates the inherent trade-off bottleneck between spatial detail enhancement and spectral fidelity preservation. In terms of architectural design, the MSG module bridges global semantics and local details through a multi-scale pyramidal structure. Within each scale, the collaborative design of SpaCAM and SpeCAM achieves complementary modeling of spatial contextual information and long-range spectral dependencies. Furthermore, the SSCFM enables adaptive feature alignment and dynamic interaction, effectively suppressing the interference of redundant information while enhancing cross-modal consistency. Extensive experimental results on multiple benchmark datasets demonstrate that CoFusion consistently outperforms existing state-of-the-art methods in both spatial reconstruction and spectral fidelity. This fully validates the effectiveness, robustness, and computational efficiency advantages of the proposed collaborative framework in addressing complex image degradation and cross-domain generalization tasks.

\bibliography{main}

@ARTICLE{class_03,
  author={Cao, Xianghai and Yu, Jiayu and Xu, Ruijie and Wei, Jiaxuan and Jiao, Licheng},
  journal={IEEE Transactions on Geoscience and Remote Sensing}, 
  title={Mask-Enhanced Contrastive Learning for Hyperspectral Image Classification}, 
  year={2024},
  volume={62},
  number={},
  pages={1-15},
  doi={10.1109/TGRS.2024.3479220}
}

@ARTICLE{class_04,
  author={Bai, Yu and Wu, Haoqi and Zhang, Lili and Guo, Hanlin},
  journal={IEEE Geoscience and Remote Sensing Letters}, 
  title={Lightweight Mamba Model Based on Spiral Scanning Mechanism for Hyperspectral Image Classification}, 
  year={2025},
  volume={22},
  number={},
  pages={1-5},
  doi={10.1109/LGRS.2025.3543315}}

@inproceedings{HRNet,
	title={Deep high-resolution representation learning for human pose estimation},
	author={Sun, Ke and Xiao, Bin and Liu, Dong and Wang, Jingdong},
	booktitle={Proceedings of the IEEE/CVF conference on computer vision and pattern recognition},
	pages={5693--5703},
	year={2019}
}

@ARTICLE{MHFnet,
	author={Xie, Qi and Zhou, Minghao and Zhao, Qian and Xu, Zongben and Meng, Deyu},
	journal={IEEE Transactions on Pattern Analysis and Machine Intelligence},
	title={MHF-Net: An Interpretable Deep Network for Multispectral and Hyperspectral Image Fusion},
	year={2022},
	volume={44},
	number={3},
	pages={1457-1473},
	doi={10.1109/TPAMI.2020.3015691}}

@article{HSRnet,
 title={Hyperspectral Image Super-Resolution via Deep Spatiospectral Attention Convolutional Neural Networks},
 DOI={10.1109/tnnls.2021.3084682},
 journal={IEEE Transactions on Neural Networks and Learning Systems},
 author={Hu, Jin-Fan and Huang, Ting-Zhu and Deng, Liang-Jian and Jiang, Tai-Xiang and Vivone, Gemine and Chanussot, Jocelyn},
 year={2022},
 pages={7251–7265},
 }

@article{Chikusei,
  title={Airborne hyperspectral data over Chikusei},
  author={Yokoya, Naoto and Iwasaki, Akira},
  journal={Space Appl. Lab., Univ. Tokyo, Tokyo, Japan, Tech. Rep. SAL-2016-05-27},
  pages={5},
  year={2016}
}

@ARTICLE{Fusformer,
  author={Hu, Jin-Fan and Huang, Ting-Zhu and Deng, Liang-Jian and Dou, Hong-Xia and Hong, Danfeng and Vivone, Gemine},
  journal={IEEE Geoscience and Remote Sensing Letters},
  title={Fusformer: A Transformer-Based Fusion Network for Hyperspectral Image Super-Resolution},
  year={2022},
  pages={1-5}
}

@inproceedings{U2Net,
author = {Peng, Siran and Guo, Chenhao and Wu, Xiao and Deng, Liang-Jian},
title = {U2Net: A General Framework with Spatial-Spectral-Integrated Double U-Net for Image Fusion},
year = {2023},
isbn = {9798400701085},
doi = {10.1145/3581783.3612084},
booktitle = {Proceedings of the 31st ACM International Conference on Multimedia},
pages = {3219–3227},
}

@ARTICLE{PSRT,
  author={Deng, Shang-Qi and Deng, Liang-Jian and Wu, Xiao and Ran, Ran and Hong, Danfeng and Vivone, Gemine},
  journal={IEEE Transactions on Geoscience and Remote Sensing},
  title={PSRT: Pyramid Shuffle-and-Reshuffle Transformer for Multispectral and Hyperspectral Image Fusion},
  year={2023},
  pages={1-15}
  }

@INPROCEEDINGS{SENet,
  author={Hu, Jie and Shen, Li and Sun, Gang},
  booktitle={2018 IEEE/CVF Conference on Computer Vision and Pattern Recognition},
  title={Squeeze-and-Excitation Networks},
  year={2018},
  pages={7132-7141}
}

@article{agriculture,
  title={Hyperspectral imagery applications for precision agriculture-a systemic survey},
  author={Sethy, Prabira Kumar and Pandey, Chanki and Sahu, Yogesh Kumar and Behera, Santi Kumari},
  journal={Multimedia Tools and Applications},
  volume={81},
  pages={3005--3038},
  year={2022},
}

@inproceedings{class_01,
  title={Dual-Stage Hyperspectral Image Classification Model with Spectral Supertoken},
  author={Liu, Peifu and Xu, Tingfa and Wang, Jie and Chen, Huan and Bai, Huiyan and Li, Jianan},
  booktitle={European Conference on Computer Vision},
  pages={368--386},
  year={2025},
  organization={Springer}
}

@article{class_02,
 title={Hyperspectral Image Classification With Attention-Aided CNNs},
 journal={IEEE Transactions on Geoscience and Remote Sensing},
 author={Hang, Renlong and Li, Zhu and Liu, Qingshan and Ghamisi, Pedram and Bhattacharyya, Shuvra S.},
 year={2021},
 month={Mar},
 pages={2281–2293},
 }

@article{SSIM,
 title={Image Quality Assessment: From Error Visibility to Structural Similarity},
 url={http://dx.doi.org/10.1109/tip.2003.819861},
 DOI={10.1109/tip.2003.819861},
 journal={IEEE Transactions on Image Processing},
 author={Wang, Z. and Bovik, A.C. and Sheikh, H.R. and Simoncelli, E.P.},
 year={2004},
 pages={600–612},
 }

@article{SMGU-Net,
title = {Spatial–spectral unfolding network with mutual guidance for multispectral and hyperspectral image fusion},
journal = {Pattern Recognition},
volume = {161},
pages = {111277},
year = {2025},
issn = {0031-3203},
doi = {https://doi.org/10.1016/j.patcog.2024.111277},
author = {Jun Yan and Kai Zhang and Qinzhu Sun and Chiru Ge and Wenbo Wan and Jiande Sun and Huaxiang Zhang},
}

@article{3DT-Net,
title = {Learning a 3D-CNN and Transformer prior for hyperspectral image super-resolution},
journal = {Information Fusion},
volume = {100},
pages = {101907},
year = {2023},
issn = {1566-2535},
doi = {https://doi.org/10.1016/j.inffus.2023.101907},
author = {Qing Ma and Junjun Jiang and Xianming Liu and Jiayi Ma},
}

@inproceedings{lka1,
  title={Internimage: Exploring large-scale vision foundation models with deformable convolutions},
  author={Wang, Wenhai and Dai, Jifeng and Chen, Zhe and Huang, Zhenhang and Li, Zhiqi and Zhu, Xizhou and Hu, Xiaowei and Lu, Tong and Lu, Lewei and Li, Hongsheng and others},
  booktitle={Proceedings of the IEEE/CVF conference on computer vision and pattern recognition},
  pages={14408--14419},
  year={2023}
}

@Article{lka2,
  author  = {Zhuang Liu and Hanzi Mao and Chao-Yuan Wu and Christoph Feichtenhofer and Trevor Darrell and Saining Xie},
  title   = {A ConvNet for the 2020s},
  journal = {Proceedings of the IEEE/CVF Conference on Computer Vision and Pattern Recognition (CVPR)},
  year    = {2022},
}

@InProceedings{lka3,
author = {Peng, Chao and Zhang, Xiangyu and Yu, Gang and Luo, Guiming and Sun, Jian},
title = {Large Kernel Matters -- Improve Semantic Segmentation by Global Convolutional Network},
booktitle = {Proceedings of the IEEE Conference on Computer Vision and Pattern Recognition (CVPR)},
month = {July},
year = {2017}
}

@inproceedings{ESSAformer,
  title={ESSAformer: Efficient transformer for hyperspectral image super-resolution},
  author={Zhang, Mingjin and Zhang, Chi and Zhang, Qiming and Guo, Jie and Gao, Xinbo and Zhang, Jing},
  booktitle={Proceedings of the IEEE/CVF International Conference on Computer Vision},
  pages={23073--23084},
  year={2023}
}

@inproceedings{FMPM-DNet,
  title={FMPM-DNet: Hyperspectral Pansharpening Dynamic Network Based on Feature Modulation and Probability Mask},
  author={Wang, Xiaozheng and Yang, Yong and Huang, Shuying and Lu, Hangyuan and Wan, Weiguo and Zhao, Aoqi},
  booktitle={Proceedings of the AAAI Conference on Artificial Intelligence},
  volume={39},
  pages={861--868},
  year={2025}
}

@article{BUGPan,
  title={Bidomain uncertainty gated recursive network for pan-sharpening},
  author={Hou, Junming and Liu, Xinyang and Wu, Chenxu and Cong, Xiaofeng and Huang, Chentong and Deng, Liang-Jian and You, Jian Wei},
  journal={Information Fusion},
  volume={118},
  pages={102938},
  year={2025},
  publisher={Elsevier}
}

@inproceedings{3,
  title={Toju Duke, Lucas Dixon, Kun Zhang, Quoc Le, Yonghui Wu, Zhifeng Chen, and Claire Cui. GLaM: Efficient scaling of language models with mixture-of-experts},
  author={Du, Nan and Huang, Yanping and Dai, Andrew M and Tong, Simon and Lepikhin, Dmitry and Xu, Yuanzhong and Krikun, Maxim and Zhou, Yanqi and Yu, Adams Wei and Firat, Orhan and others},
  booktitle={Proceedings of the 39th International Conference on Machine Learning},
  volume={162},
  pages={5547--5569},
  year={2022}
}

@article{PCA,
  title={Extracting spectral contrast in Landsat Thematic Mapper image data using selective principal component analysis},
  author={Kwarteng, P and Chavez, A},
  journal={Photogramm. Eng. Remote Sens},
  volume={55},
  number={1},
  pages={339--348},
  year={1989}
}

@misc{laben2000process,
  title={Process for enhancing the spatial resolution of multispectral imagery using pan-sharpening},
  author={Laben, Craig A and Brower, Bernard V},
  year={2000},
  month=jan # "~4",
  publisher={Google Patents},
  note={US Patent 6,011,875}
}

@article{BDSD,
  title={Optimal MMSE pan sharpening of very high resolution multispectral images},
  author={Garzelli, Andrea and Nencini, Filippo and Capobianco, Luca},
  journal={IEEE Transactions on Geoscience and Remote Sensing},
  volume={46},
  number={1},
  pages={228--236},
  year={2007},
  publisher={IEEE}
}

@article{SFIM,
  title={Smoothing filter-based intensity modulation: A spectral preserve image fusion technique for improving spatial details},
  author={Liu, JG},
  journal={International Journal of remote sensing},
  volume={21},
  number={18},
  pages={3461--3472},
  year={2000},
  publisher={Taylor \& Francis}
}

@article{MTF-GLP,
  title={Intersensor statistical matching for pansharpening: Theoretical issues and practical solutions},
  author={Alparone, Luciano and Garzelli, Andrea and Vivone, Gemine},
  journal={IEEE Transactions on Geoscience and Remote Sensing},
  volume={55},
  number={8},
  pages={4682--4695},
  year={2017},
  publisher={IEEE}
}

@article{AWLP,
  title={Introduction of sensor spectral response into image fusion methods. Application to wavelet-based methods},
  author={Otazu, Xavier and Gonz{\'a}lez-Aud{\'\i}cana, Mar{\'\i}a and Fors, Octavi and N{\'u}{\~n}ez, Jorge},
  journal={IEEE Transactions on Geoscience and Remote Sensing},
  volume={43},
  number={10},
  pages={2376--2385},
  year={2005},
  publisher={IEEE}
}

@article{DWT,
  title={A theory for multiresolution signal decomposition: the wavelet representation},
  author={Mallat, Stephane G},
  journal={IEEE transactions on pattern analysis and machine intelligence},
  volume={11},
  number={7},
  pages={674--693},
  year={2002},
  publisher={Ieee}
}

@article{vo_01,
  title={A new pansharpening algorithm based on total variation},
  author={Palsson, Frosti and Sveinsson, Johannes R and Ulfarsson, Magnus O},
  journal={IEEE Geoscience and Remote Sensing Letters},
  volume={11},
  number={1},
  pages={318--322},
  year={2013},
  publisher={IEEE}
}

@article{vo_02,
  title={A variational pansharpening approach based on reproducible kernel Hilbert space and heaviside function},
  author={Deng, Liang-Jian and Vivone, Gemine and Guo, Weihong and Dalla Mura, Mauro and Chanussot, Jocelyn},
  journal={IEEE Transactions on Image Processing},
  volume={27},
  number={9},
  pages={4330--4344},
  year={2018},
  publisher={IEEE}
}

@article{MRA_01,
    title = {A theoretical and practical survey of image fusion methods for multispectral pansharpening},
    journal = {Information Fusion},
    volume = {79},
    pages = {1-43},
    year = {2022},
    issn = {1566-2535},
    doi = {https://doi.org/10.1016/j.inffus.2021.10.001},
    url = {https://www.sciencedirect.com/science/article/pii/S1566253521001998},
    author = {Cigdem Serifoglu Yilmaz and Volkan Yilmaz and Oguz Gungor},

}

@article{MRA_02,
  title={A Metaheuristic Optimization-Based Solution to MTF-GLP-Based Pansharpening},
  author={Serifoglu Yilmaz, Cigdem and Gungor, Oguz},
  journal={PFG--Journal of Photogrammetry, Remote Sensing and Geoinformation Science},
  volume={91},
  number={4},
  pages={245--272},
  year={2023},
  publisher={Springer}
}

@inproceedings{MRA_03,
  title={Multi resolution analysis (mra) for approximate self-attention},
  author={Zeng, Zhanpeng and Pal, Sourav and Kline, Jeffery and Fung, Glenn M and Singh, Vikas},
  booktitle={International conference on machine learning},
  pages={25955--25972},
  year={2022},
  organization={PMLR}
}

@article{zhu2024implicit,
  title={An implicit transformer-based fusion method for hyperspectral and multispectral remote sensing image},
  author={Zhu, Chunyu and Zhang, Tinghao and Wu, Qiong and Li, Yachao and Zhong, Qin},
  journal={International Journal of Applied Earth Observation and Geoinformation},
  volume={131},
  pages={103955},
  year={2024},
  publisher={Elsevier}
}

@inproceedings{hu2019local,
  title={Local Relation Networks for Image Recognition},
  author={Hu, Han and Zhang, Zheng and Xie, Zhenda and Lin, Stephen},
  booktitle={Proceedings of the IEEE/CVF International Conference on Computer Vision (ICCV)},
  pages={3464--3473},
  year={2019}
}

@inproceedings{liu2021Swin,
  title={Swin Transformer: Hierarchical Vision Transformer using Shifted Windows},
  author={Liu, Ze and Lin, Yutong and Cao, Yue and Hu, Han and Wei, Yixuan and Zhang, Zheng and Lin, Stephen and Guo, Baining},
  booktitle={Proceedings of the IEEE/CVF International Conference on Computer Vision (ICCV)},
  year={2021}
}

@inproceedings{liu2021swinv2,
  title={Swin Transformer V2: Scaling Up Capacity and Resolution}, 
  author={Ze Liu and Han Hu and Yutong Lin and Zhuliang Yao and Zhenda Xie and Yixuan Wei and Jia Ning and Yue Cao and Zheng Zhang and Li Dong and Furu Wei and Baining Guo},
  booktitle={International Conference on Computer Vision and Pattern Recognition (CVPR)},
  year={2022}
}

@InProceedings{derconv_01,
  author    = {Dai, Jifeng and Qi, Haozhi and Xiong, Yuwen and Li, Yi and Zhang, Guodong and Hu, Han and Wei, Yichen},
  booktitle = {Proceedings of the IEEE international conference on computer vision},
  title     = {Deformable convolutional networks},
  pages     = {764--773},
  year      = {2017},
}

@InProceedings{derconv_02,
  author    = {Zhu, Xizhou and Hu, Han and Lin, Stephen and Dai, Jifeng},
  booktitle = {Proceedings of the IEEE/CVF conference on computer vision and pattern recognition},
  title     = {Deformable convnets v2: More deformable, better results},
  pages     = {9308--9316},
  year      = {2019},
}

@inproceedings{li2026pif,
  title={PIF-Net: Ill-Posed Prior Guided Multispectral and Hyperspectral Image Fusion via Invertible Mamba and Fusion-Aware LoRA},
  author={Li, Baisong and Wang, Xingwang and Xu, Haixiao},
  booktitle={Proceedings of the AAAI Conference on Artificial Intelligence},
  volume={40},
  number={8},
  pages={5964--5972},
  year={2026}
}

\end{document}